\documentclass[10pt,twocolumn,letterpaper]{article}

\usepackage{iccv}
\usepackage{times}
\usepackage{epsfig}
\usepackage{graphicx}
\usepackage{amsmath} 
\usepackage{amssymb}
\usepackage{booktabs}
\usepackage{multirow}
 \usepackage{amssymb}

\usepackage[pagebackref=true,breaklinks=true,letterpaper=true,colorlinks,bookmarks=false]{hyperref}

\iccvfinalcopy 

\def\iccvPaperID{****} 

\ificcvfinal\fi

\begin{document}

\title{Learning to Disentangle GAN Fingerprint for Fake Image Attribution}

\author{Tianyun Yang$^{12}$, Juan Cao$^{12*}$, Qiang Sheng$^{12}$, Lei Li$^{12}$, Jiaqi Ji$^{3}$, Xirong Li$^{3}$, Sheng Tang$^{12}$\\
$^1$
Institute of Computing Technology, Chinese Academy of Sciences, Beijing, China\\
$^2$University of Chinese Academy of Sciences, Beijing, China\\
$^3$Renmin University of China\\
{\tt\small \{yangtianyun19z,caojuan,shengqiang18z,lilei17b,ts\}@ict.ac.cn}\\
{\tt\small \{2019104238,xirong\}@ruc.edu.cn}
}

\maketitle
\ificcvfinal\thispagestyle{empty}\fi

\begin{abstract}
Rapid pace of generative models has brought about new threats to visual forensics such as malicious personation and digital copyright infringement, which promotes works on  fake image attribution. Existing works on fake image attribution mainly rely on a direct classification framework.
Without additional supervision, the extracted features could include many content-relevant components and generalize poorly. Meanwhile, how to obtain an interpretable GAN fingerprint to explain the decision remains an open question. Adopting a multi-task framework, we propose a GAN Fingerprint Disentangling Network (GFD-Net) to simultaneously disentangle the fingerprint from GAN-generated images and produce a content-irrelevant representation for fake image attribution. A series of constraints are provided to guarantee the   stability and discriminability of the fingerprint, which in turn helps content-irrelevant feature extraction. Further, we perform comprehensive analysis on GAN fingerprint, providing some clues about the properties of GAN fingerprint and which factors dominate the fingerprint in GAN architecture. Experiments show that our GFD-Net achieves superior fake image attribution performance in both closed-world and open-world testing. We also apply our method in binary fake image detection and exhibit a significant generalization ability on unseen generators. 
\end{abstract}

\section{Introduction}

The progressive generation technology has produced extremely realistic generated images, 
	which raises big challenges to visual forensics. Dedicated
	research efforts are paid \cite{durall2020watch,wang2020cnn,liu2020global,zhang2019detecting,
	jeon2020t,nataraj2019detecting,chai2020makes,frank2020leveraging} to detect generated images in recent years. 
	However, only real/fake classification is not the end: For malicious and illegal content, law enforcers need to identify its owner;
	For GAN developers, GAN models needs experienced experts to design with laborious trial-and-error testings and some have high commercial value, which should be protected. To these ends, we aim at the task of fake image attribution, i.e., attributing the origin of fake images.

\begin{figure}[t]
\begin{center}
\includegraphics[width=\linewidth]{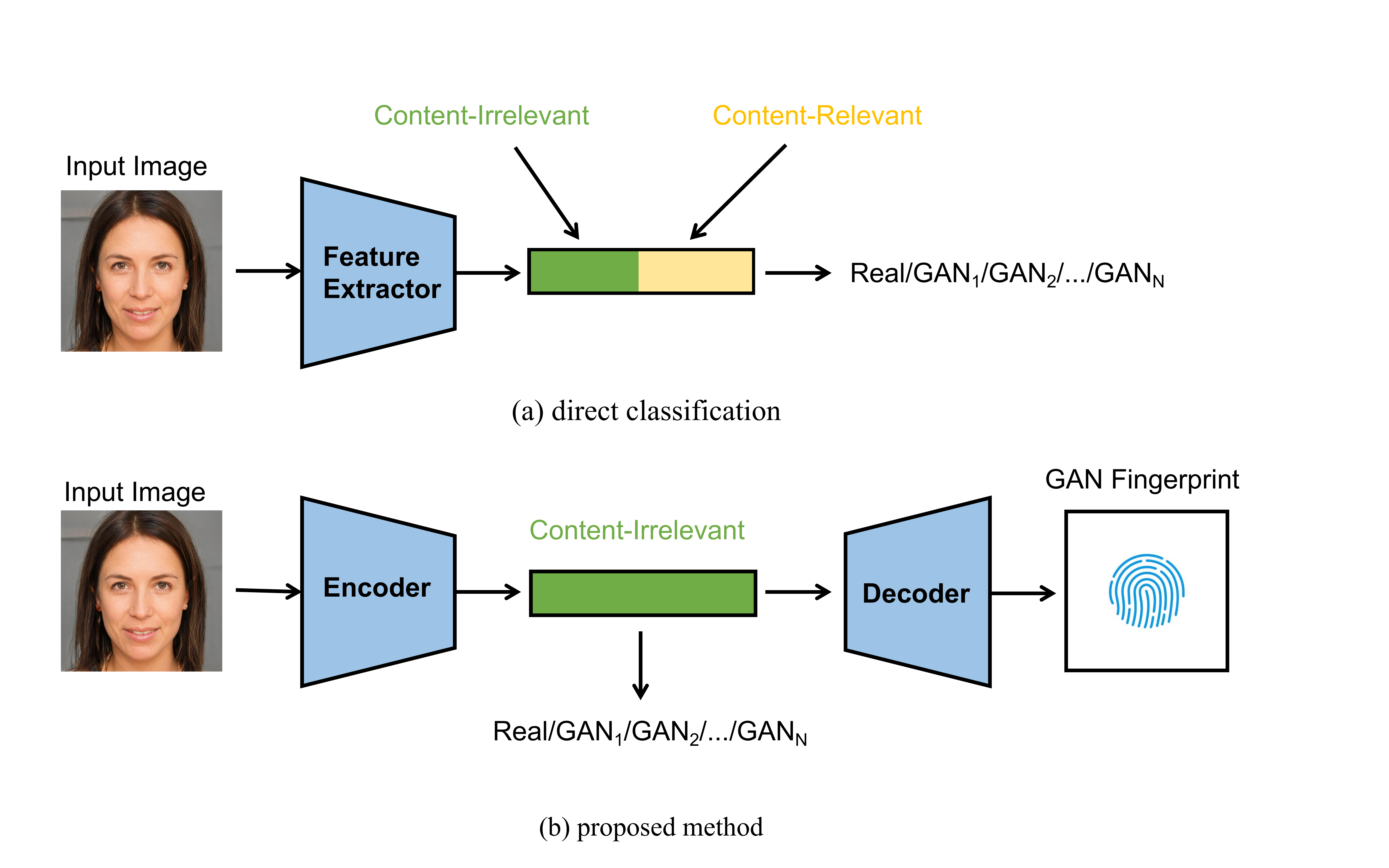} 
\end{center}
   \caption{Comparison between (a) direct classification method and (b) our method for fake image attribution. Features extracted by direct classification unavoidably contain content-relevant components, while our method disentangles GAN fingerprint from input images, and simultaneously produce a content-irrelevant representation for fake image attribution.}  

\label{fig:example}
\end{figure}

In previous works, Marra \etal \cite{marra2019gans} take averaged noise image as the GAN fingerprint, showing each GAN leaves its specific fingerprint on the images it generates. 
Yu \etal \cite{yu2019attributing} decouple the GAN fingerprint into model fingerprint and image fingerprint. Specifically, they take the model's output feature as image fingerprint and the parameters in the last layer as model fingerprint. Then attribution is achieved by the interaction between image and model fingerprint. Frank \etal \cite{frank2020leveraging} leverage a frequency-domain view and take the discrete cosine transform (DCT) transformed image as classifier's input for source identification. 

While encouraging, there are two problems in existing works:
First, how to visualize GAN fingerprint remains an open question. Although Marra \etal \cite{marra2019gans} and Yu \etal \cite{yu2019attributing} propose to visualize the fingerprint by averaged noise residual and auto-encoder reconstruction residual, the visualized fingerprints still contain many redundant noise, which makes it hard to observe the difference between different GANs and the shared properties in the images from the same GAN. Therefore, we aim at generating GAN fingerprints that are {\em common} and {\em stable} among images generated by the same GAN and {\em distinct} between different GANs.

Second, without additional supervision, direct classification methods would harvest any useful features to help classification, which may include many content-relevant information such as explicit artifacts as shown in Figure~\ref{fig:example}(a). 
However, a qualified GAN fingerprint should remain stable no matter what content the GAN generates. Although the learned representation by classifier-based method is discriminative enough to handle seen images, it may generalize poorly on images with different content. Therefore, we intend to make the model focus on {\em content-irrelevant} features.

In this paper, we propose a GAN Fingerprint Disentangling Network (GFD-Net) (Figure~\ref{fig:example}b). GFD-Net has two goals: 1) disentangle fingerprints from GAN-generated images; 2) produce a content-irrelevant representation for fake image attribution. The two goals are achieved by the joint learning among a generator $G$, a discriminator $D$ and an auxiliary classifier $C$. $G$ serves as a fingerprint extractor,  we overlap the extracted fingerprint from $G$ with a real image to obtain a fingerprinted image. $D$ and $C$ make the learned fingerprint content-irrelevant and discriminative by supervising the fingerprinted image. 
The fingerprint learning in turn helps $G$ to extract content-irrelevant features that represent a certain GAN specifically. Thus we use the bottleneck feature of $G$  for fake image attribution.
 
To demonstrate the effectiveness of GFD-Net, we conduct cross-dataset fake image attribution experiments and apply our method on cross-generator fake image detection. Extensive experiments demonstrate the superior generalization ability of our method. 

With the disentangled fingerprints, we further investigate the properties of GAN fingerprint and qualitatively analyze which factors in GAN architecture dominate the fingerprint. We show that GAN fingerprint is mostly influenced by the construction and combination of layers, while changing feature channel number has less effect on it.
To summarize, the contributions of this work include:

\begin{itemize}
\setlength{\itemsep}{0pt}
\setlength{\parsep}{0pt}
\setlength{\parskip}{0pt}
\item We propose a GAN Fingerprint Disentangling Network (GFD-Net), which can disentangle the fingerprint from GAN-generated images and simultaneously produce content-irrelevant representation for fake image attribution.

\item We successfully extract GAN fingerprints that are common and stable among images generated by the same GAN and distinct between different GANs. With the learned fingerprint, we investigate the properties of GAN fingerprints and qualitatively analyze how GAN architecture dominates GAN fingerprint.

\item Extensive experiments demonstrate that GFD-Net has superior generalization ability in not only fake image attribution but also fake image detection.

\end{itemize}

\section{Related Work}

\noindent {\bf Fake image detection.}
	Along with the rapid development of generation technology, concerns are raised about the malicious use of generated images. Some researchers have paid effort to address the problem
	of fake image detection~\cite{durall2020watch,wang2020cnn,liu2020global,zhang2019detecting,
	jeon2020t,nataraj2019detecting,chai2020makes,frank2020leveraging}. Among these works, the generalization ability of the detection method has been paid close attention to.
	Some works~\cite{durall2020watch,frank2020leveraging,zhang2019detecting} exploited the common checkerboard artifacts caused by upsampling operation in GAN architecture, and model this artifact in the frequency domain. Liu \etal~\cite{liu2020global} analyzed texture statistics of fake images and
	adopted the Gram matrix to capture global or long-range texture for better generalization
	ability. Wang \etal~\cite{wang2020cnn} experimented on images created from a variety of CNN models and revealed that there exist common artifacts generalized from one model to another.
	Jeon \etal~\cite{jeon2020t} designed a transferable framework to improve the transferability of GAN image detection. Chai \etal~\cite{chai2020makes} proposed to use classifiers with limited receptive fields to focus on local common artifacts shared by different GANs. However, these works explore little on the inherent difference between images from different GANs. We propose a novel network to disentangle the interpretable fingerprint for each GAN. \\
	\noindent {\bf Fake image attribution.} Fake image attribution can be classified into passive attribution~\cite{kim2020decentralized,yu2020artificial, yu2020responsible} and positive attribution~\cite{marra2019gans,yu2019attributing, yu2019attributing}. Works on positive attribution insert artificial fingerprint~\cite{yu2020artificial,yu2020responsible} or inject key~\cite{kim2020decentralized} directly to the generation model and then decouple the fingerprint or key when tracing the source model.
	Compared with positive attribution, passive attributing is more challenging and applicable. Marra \etal~\cite{marra2019gans} find averaged noise residual can represent the GAN fingerprint. Frank \etal~\cite{frank2020leveraging} observe the discrepant DCT frequency spectrums exhibited by images generated from different GAN architectures, and then send the DCT frequency spectrum into classifiers for source identification.
  Yu \etal~\cite{yu2019attributing} decouple GAN fingerprint into model fingerprint and image fingerprint. Specifically, they take the model’s output feature as image fingerprint and the parameters in the last layer as model fingerprint
	 Then attribution is achieved by the interaction between model and image fingerprint. However, the extracted fingerprints by these works tend to contain content-relevant information and thus lack generalization. In our work, we aim at using a learning-based method to 
disentangle content-irrelevant features from the input image for fake image attribution.

\begin{figure*}
\begin{center} 
\includegraphics[width=0.7\textwidth]{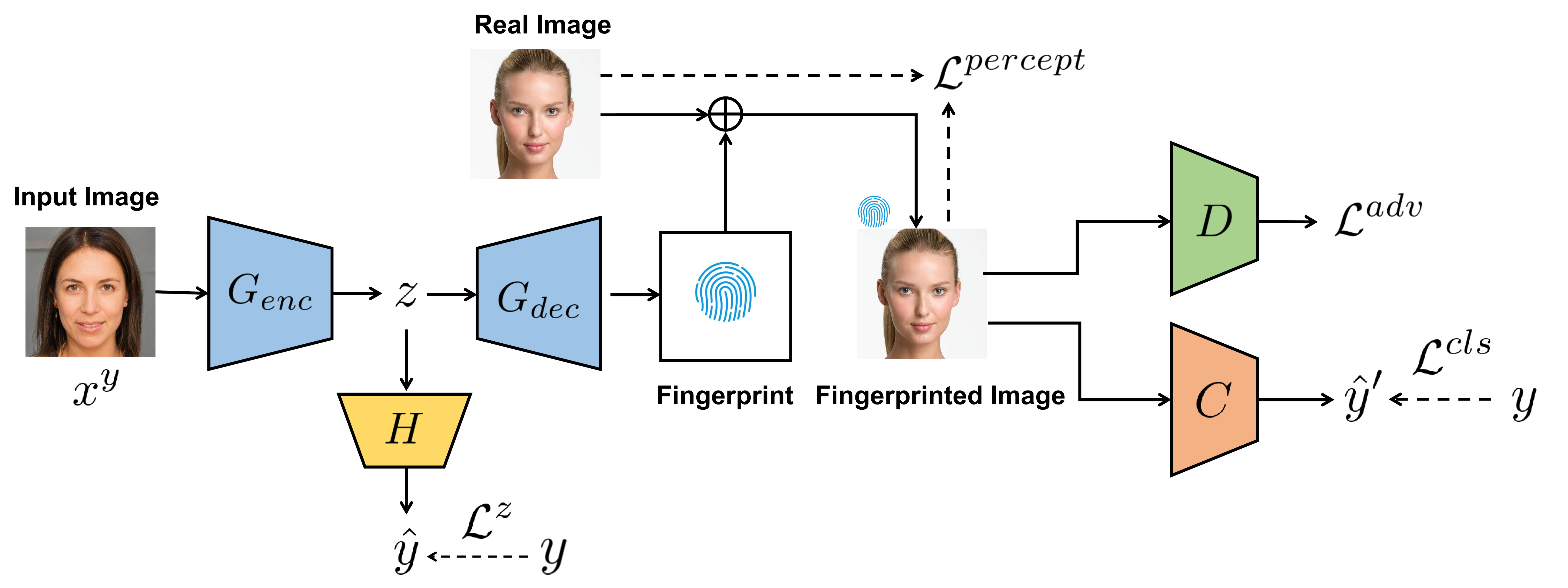}  
\end{center}
   \caption{Overall framework of GFD-Net.  (1) The generator $G$ takes in an image $x^{y}$ with source $y$ as input and outputs a fingerprint, then the fingerprint is added on a real image to composite a fingerprinted image.
(2) The auxiliary classifier $C$ supervises the fingerprinted image to be classified as the class of the input image, and thus forces the generator $G$ to generate fingerprint with discriminative properties.
   (3) The discriminator $D$ forces the fingerprinted image to be realistic, and thus demands $G$ to generate visually imperceptible and content-irrelevant fingerprint. 
 (4) A perceptual loss is applied on the fingerprinted image and corresponding real image to further suppress content-irrelevant clues on the learned fingerprint.    
   (5) The fingerprint learning process helps the encoder $G_{enc}$ to learn a representation which is both content-irrelevant and discriminative. Then a classification head $H$ is added on the latent code $z$ for fake image attribution.} 
\label{fig:framework}
\end{figure*}

\section{Proposed Method}

\subsection{Problem Formulation}

Given an image $x^{y}$ with source $y\in\mathbb{Y}=\{real,GAN_{1},GAN_{2},\dots,GAN_{N}\}$, where $GAN_{1},\dots,$ $GAN_{N}$ have different architectures. The goal of image attribution is to learn a mapping $D(x^{y}) \rightarrow y$~\cite{yu2020artificial}. There are two goals in our learning process: 1) to learn a {\em content-irrelevant}  feature representation for fake image attribution, and 2) to visualize GAN fingerprints that are common and stable among images generated by the same GAN and distinct between different GANs.

\subsection{Network Structure}

Figure~\ref{fig:framework} shows the overall architecture of GFD-Net. The network adopts a GAN-like framework, which comprises a generator $G$, a discriminator $D$, and an auxiliary classifier $C$. 
The fingerprint generator $G$ uses a U-Net~\cite{ronneberger2015u} structure with skip connections from the encoder $G_{enc}$ to the decoder $G_{dec}$. $G_{enc}$  projects the input image  $x^{y}$ into a latent vector $z$, and then $G_{dec}$ transforms $x^{y}$ into a fingerprint $f$ with the same size as the input image. 

Unlike conventional generators, we add a classification head $H$ on $z$ to make the learned feature more discriminative. The classification head comprises an average pooling layer and a fully-connected layer, which takes the latent code $z$ as input and outputs the source prediction $\hat{y}$.

After getting the fingerprint, we add it on a randomly selected real image $x^{real}$ to generate a fingerprinted image $x^{fp}$:
\begin{equation}
\begin{aligned}
x^{fp} &=  x^{y}+ f
\end{aligned}
\end{equation}
Then the fingerprinted image $x^{fp}$ is sent to a discriminator $D$ and an auxiliary classifier $C$. 
 
 For the discriminator $D$, we use a PatchGAN architecture with 3 convolutional layers following the implementation in ~\cite{isola2017image}. The objective of $D$ is to classify the input image $x$ as real and  fingerprint image $x^{fp}$ as fake. For the auxiliary classifier $C$, we use a ResNet-50~\cite{he2016deep} architecture. $C$ is trained to predict the source of an image, which 
aims at making the fingerprinted image classified as the same class as the input image.
 
\subsection {Loss Functions} 
Having defined the overall structure, we now move on to discuss how we formulate our objective for learning.

\noindent{\bf Auxiliary Classification Loss}.  The auxiliary classification loss is added on the auxiliary classifier, which is proposed to make the learned fingerprint distinct between different GANs. Based on a prior that if the learned fingerprint is representative of its class, then when it is added on a real image, the fingerprinted image should own similar properties with the input image that generate the fingerprint. Thus, we employ an auxiliary classification loss on the fingerprinted image $x^{fp}$ and constrain it to be classified as the same class as the input image (i.e., $C(x^{fp}) \rightarrow y$) by minimizing
\begin{equation}
\begin{aligned}
\mathcal{L}_{G}^{cls} &= \mathcal{L}_{CE} (C(x^{fp}),y)
\end{aligned}
\end{equation}

The auxiliary classifier $C$ is trained previously on the input images with multiples source by minimizing
\begin{equation}
\begin{aligned}
\mathcal{L}_{C}^{cls} &= \mathcal{L}_{CE}(C(x^{y}),y)
\end{aligned}
\end{equation}

 \noindent{\bf Adversarial Loss}. 
 The  auxiliary classification loss is proposed to make the learned fingerprint representative of its class. However, with only an auxiliary classification loss, the learned fingerprint would still contain content-relevant information and become unstable within the same class.
  Hence, an adversarial loss is proposed to suppress the learning of content-relevant features. 
  
  The adversarial loss is introduced between the generator and discriminator aiming at making the fingerprinted image $x^{fp}$ realistic. In this way, the generated fingerprint $f$ is expected to be visually imperceptible when added to a real image, which mediately forces the generator $G$ to extract content-irrelevant features from input images. The adversarial losses for the discriminator and the generator are formulated as
\begin{equation}
\begin{aligned}
\mathcal{L}_{D}^{adv} &=\mathbb{E}[\log(1-D(x^{fp}))] +\mathbb{E}[\log (D(x)] \\
\mathcal{L}_{G}^{adv} &=\mathbb{E}[\log (D(x^{fp}))] \\
\end{aligned}
\end{equation}

When training $D$ by minimizing $L_{D}^{adv}$, $D$ is encouraged to distinguish between fingerprinted images and real GAN images (the input images). When training the generator $G$, the fingerprinted images are expected to fool $D$. With $D$ as a supervision, the generator learns to extract stable content-irrelevant fingerprints from input images.

\noindent{\bf Perceptual Loss}. To further make the fingerprinted image visually consistent with the real image and restrain the content-relevant information on the fingerprinted image, we adopt a VGG-16 perceptual loss~\cite{johnson2016perceptual} between fingerprint image and corresponding real image.
\begin{equation}
\begin{aligned}
\mathcal{L}^{percept}_{G} &= \| F(x^{fp})-F(x^{real}) \|_{2}
\end{aligned}
\end{equation}
where $\| \cdot \|_{2}$ denotes $l_{2}$ distance, $F$ denotes a VGG feature extraction model. 

\noindent{\bf Latent Classification Loss.} The latent classification loss is added on the classification head, which has two functions: 1) It makes the encoder learn discriminative feature of each class and helps the generation of representative fingerprint. 2) The fingerprint learning process in turn helps the encoder to produce a content-irrelevant representation, then the latent classification loss is optimized to map the latent code $z$ to the source $y$ for fake image attribution and is formulated as
\begin{equation}
\begin{aligned}
\mathcal{L}_{G}^{z} &= \mathcal{L}_{CE}(H(z),y)
\end{aligned}
\end{equation}
where $\mathcal{L}_{CE}$ is the cross entropy loss for classification.

\subsection{Overall Objective} 
Combining all components described above, our two objectives are achieved: 1) The generator $G$ takes an image as input and outputs the fingerprint corresponding to its source. 2) Benefit from fingerprint learning, the encoder $G_{enc}$ produces a content-irrelevant and discriminative representation which facilitates fake image attribution. The classification head $H$ attributes the input image to its source.

The training process contains two steps:
In the first step, we train generator $G$ with $D$ and $C$ fixed. In the next step, we keep $G$ fixed and train $D$ and $C$. Overall, the objective for the generator (include the classification head) is formulated as
\begin{equation}
\begin{aligned}
\mathcal{L}_{G} &= \omega_{1}\mathcal{L}^{z}_{G}+\omega_{2}\mathcal{L}^{adv}_{G}+\omega_{3}\mathcal{L}^{cls}_{G}+\omega_{4}\mathcal{L}^{percept}_{G}
\end{aligned}
\end{equation}
and the objective for the discriminator and the auxiliary classifier is formulated as
\begin{equation}
\begin{aligned}
\mathcal{L}_{D,C} &= \mathcal{L}^{cls}_{C}+\mathcal{L}^{adv}_{D}
\end{aligned}
\end{equation}
where $\omega_{i} (i=1,\dots,4)$ are non-negative weights.

\section{Experiments}

\subsection{Setup}
\label{sec:setup}
\noindent {\bf Baselines.} We compare GFD-Net with the following methods: 1) PRNU~\cite{marra2019gans}: a method using photo-response non-uniformity (PRNU) patterns as the fingerprint for fake image attribution. 
2) DCT~\cite{frank2020leveraging}: a frequency-based method that uses DCT transformed images for fake image attribution and detection.
3) AttNet~\cite{yu2019attributing}: a PatchGAN-like classifier for fake image attribution. 
4) CNNDetect~\cite{wang2020cnn}: a fake image detection method which uses ResNet-50 as classifier. 
5) PatchForensics~\cite{chai2020makes}: use classifiers with limited receptive fields to focus on common artifacts generalized between different GAN models.
6) Xception~\cite{chollet2017xception} and DenseNet~\cite{huang2017densely}: two widely-used CNNs for image representation.

\noindent{\bf Implementation details.} 
Adam optimizer is used with initial learning rate 1e-4. We use a step decay scheduler with gamma as 0.9 and step size as 500. We set $\omega_{1}$, $\omega_{2}$, $\omega_{3}$ and $\omega_{4}$ as 10,1e-1,1,1 for fake image attribution experiment and 10,1e-2,1,1 for fake image detection experiment. 

\noindent{\bf Datasets.} For fake image attribution in Section \ref{sec:att}, we consider following GAN architectures: ProGAN~\cite{karras2017progressive}, MMDGAN~\cite{binkowski2018demystifying}, SNGAN~\cite{miyato2018spectral} and InfoMaxGAN~\cite{lee2021infomax}, StyleGAN~\cite{karras2019style}, StyleGAN2~\cite{karras2020analyzing}. For fake image detection in Section \ref{sec:general}, we use {\em ForenSynths} dataset~\cite{wang2020cnn}, which include 13 synthesis algorithms: ProGAN, StyleGAN, StyleGAN2, whichfaceisreal(WFIR)\footnote{\url{https://www.whichfaceisreal.com/}}, BigGAN~\cite{brock2018large}, CycleGAN~\cite{zhu2017unpaired}, StarGAN~\cite{choi2018stargan}, GauGAN~\cite{park2019gaugan}, Cascaded Refinement Networkd(CRN)~\cite{chen2017photographic}, Implicit Maximun Likelihood Estimation(IMLE)~\cite{li2019diverse},  
Second Order Attention Network (SAN)~\cite{dai2019second}, Seeing In The Dark(SITD)~\cite{chen2018learning} and FaceForensics++(FF++)~\cite{rossler2019faceforensics++}.

\noindent{\bf Evaluation Metric.} We use classification accuracy to evaluate the performance.

\begin{figure*}
\begin{center}
\includegraphics[width=1\textwidth]{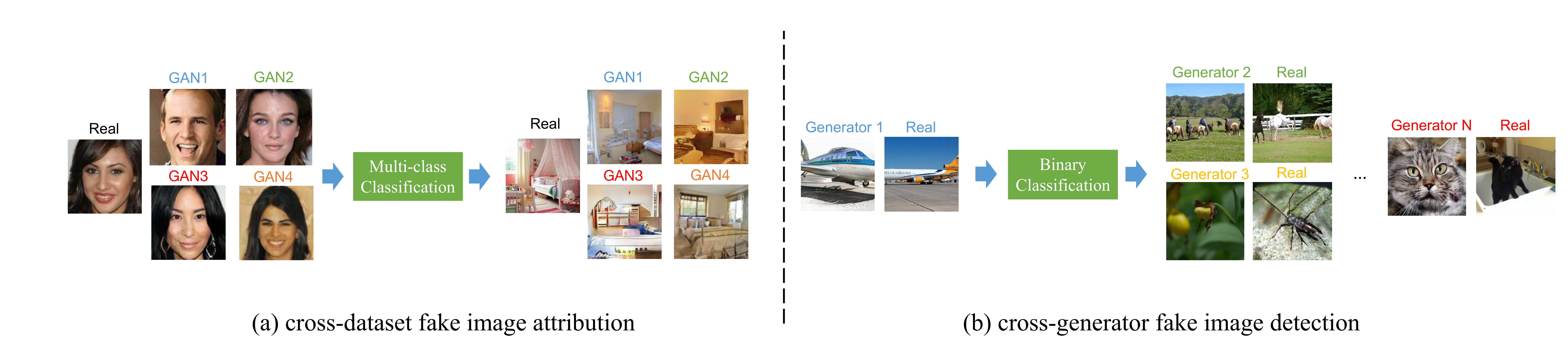} 
\end{center}
   \caption{Illustration of our two generalization experiments: (1) In cross-dataset fake image attribution experiment, we evaluate the generalization ability on fake images generated by GANs trained on different dataset. (2) In cross-generator fake image detection, we evaluate the generalization ability on images from different generators.}
\label{fig:exp} 
\end{figure*} 

\subsection{Cross-Dataset Fake Image Attribution}
\label{sec:att} 

 In this section, we conduct experiments on cross-dataset fake image attribution as illustrated in Figure \ref{fig:exp}(a). 
 To avoid the resolution influence, two experiments are implemented on 128 px and 1024 px resolution GANs respectively.  For each resolution, we test in {\bf closed world} and {\bf open world}, depending on whether or not the images in the testing set are generated by the same set of GAN models used in training. We compare with PRNU, DCT and AttNet that are originally proposed for fake image attribution.
 	
\subsubsection {Evaluation on 128px GANs}
\label{sec:128}
The experiment is conducted on 5 classes: real, ProGAN, MMDGAN, SNGAN and InfoMaxGAN. For each GAN architecture, we use two models trained respectively on CelebA dataset~\cite{liu2015deep} and LSUN bedroom  dataset~\cite{radford2015unsupervised}. We sample 20k images equally from CelebA, ProGAN-CelebA, MMDGAN-CelebA, SNGAN-CelebA and InfoMaxGAN-CelebA and then split each set into 15k training, 1k validation, and 4k for closed world testing. To evaluate the generalization capability across training dataset, we conduct open world testing on LSUN, ProGAN-LSUN, MMDGAN-LSUN, SNGAN-LSUN, and InfoMaxGAN-LSUN with 4k images for each class. -CelebA/LSUN means the model is trained on CelebA/LSUN dataset. To make the models focus on local patterns, we resize all images from 128px to 512px, then in training and testing we random and center crop the images to 224px patches.

\begin{table}[t]
\renewcommand\arraystretch{1.1}
\begin{center}
\scalebox{0.9}{
\begin{tabular}{lcccc}
\toprule
\multirow{1}{*}{Method} & \multicolumn{1}{c}{Closed-world} & &  \multicolumn{1}{c}{Open-world} \\ 
\midrule
PRNU~\cite{marra2019gans} &      92.23                 &  &  18.57                 &              \\
DCT~\cite{frank2020leveraging}&     94.40           &   &   51.26                &              \\
AttNet~\cite{yu2019attributing}  &     99.44             &   &  65.18                 &              \\
\bf GFD-Net(Ours)                    &   \bf  99.99             &   &   \bf 78.72               &            \\ \bottomrule
\end{tabular}}
\end{center}
\caption{Accuracy(\%) on 128px cross-dataset fake image attribution. The best results in each column are \bf{boldfaced}.} 
\label{tab:att128}
\end{table}

\begin{table*}
\begin{center}
 \renewcommand\arraystretch{1}
\scalebox{0.81}{
\setlength\tabcolsep{3.2pt}
\begin{tabular}{lccccccccccclcccc}
\toprule
\multirow{2}{*}{Method}  & \multirow{2}{*}{Closed world} &  \multicolumn{10}{c}{open world StyleGAN} & & \multicolumn{4}{c}{open world StyleGAN2}  \\
 \cmidrule{3-12} \cmidrule{14-17}  & & Yellow & Model & Asian star & kid  & elder & adult & glass & male & female & smile & & Yellow & Model & Asian star & kid\\
 \midrule
PRNU~\cite{marra2019gans}  &76.32 &	67.73 &	67.33 &	57.45 &	62.53 &	59.63 &	71.68 &	60.98 &	65.88&	71.08&	73.15	&& 17.63&	9.58	&6.70	&15.35 \\ 

DCT~\cite{frank2020leveraging}  & \bf 99.95 & 50.83  & 38.75 &   33.35  &  35.43   &  48.88   &   43.38  &  43.55   &   40.15  &  41.52   &   49.69  & & 30.78 & 0.20 & 0.78& 5.03 \\ 
AttNet~\cite{yu2019attributing} & 99.20 & 93.53 & 89.53 &  76.90 &  98.13   &  \bf99.08   &  99.13   & 99.48    &  \bf 99.68   &  99.05   &  98.91  & & 99.00& 0.00 &0.36&96.65\\ 
\bf GFD-Net(Ours)   &99.43&	\bf 97.85	&\bf92.40	&\bf92.30	&\bf98.80	&\bf99.08	&\bf99.23&	\bf99.50    & 99.63  & \bf99.28	&\bf99.30&&	\bf99.78	&	\bf63.95	&	\bf84.48	&	\bf99.95\\ 
\bottomrule
\end{tabular}} 
\end{center}
\caption{Accuracy(\%) on 1024px cross-dataset fake image attribution. The best results in each column are \bf{boldfaced}.} 
\label{tab:att1024}
\end{table*}

\begin{table*}
 \renewcommand\arraystretch{1.2}
\scalebox{0.71}{
\setlength\tabcolsep{3.2pt}
\begin{tabular}{@{}lccccccccccccccccccccccc@{}} 
\toprule
\multirow{3}{*}{Method} & \multicolumn{17}{c}{Test set Accuracy} & \multicolumn{1}{c}{Total}  
\\ \cmidrule(l){2-19} &
  \multicolumn{5}{c}{Unconditional GAN} & &
  \multicolumn{3}{c}{Conditional GAN} & &
  \multicolumn{2}{c}{Perceptual loss} & &
  \multicolumn{2}{c}{Low-level vision} & &
  \multicolumn{1}{c}{DeepFake} &
  \multicolumn{1}{c}{\multirow{2}{*}{\bf{\em Avg.}}} \\ 
\cmidrule(lr){2-6} \cmidrule(lr){8-10}  \cmidrule(lr){12-13} \cmidrule(lr){15-16} \cmidrule(lr){18-18}  &
  \multicolumn{1}{c}{ProGAN} &
  \multicolumn{1}{c}{StyleGAN} &
  \multicolumn{1}{c}{StyleGAN2} &
  \multicolumn{1}{c}{WFIR} &
  \multicolumn{1}{c}{BigGAN} & &
  \multicolumn{1}{c}{CycleGAN} &
  \multicolumn{1}{c}{StarGAN} &
  \multicolumn{1}{c}{GauGAN} & &
  \multicolumn{1}{c}{CRN} &
  \multicolumn{1}{c}{IMLE} & &
  \multicolumn{1}{c}{SAN} & 
  \multicolumn{1}{c}{STID} & & 
  \multicolumn{1}{c}{FF++} &
  \multicolumn{1}{c}{} \\ \midrule  
PRNU~\cite{chai2020makes}   
 &   54.03	&47.88	&48.29	&45.9&	46.85&&	48.26&	41.22 & 50.87 && 51.97	&50.92&& 51.07 &	48.06&&	50.62	& {\em 48.92}                        \\

DCT~\cite{frank2020leveraging}  & 69.06 &	78.94 &	66.94 &	67.95 &	57.58 &&	71.04 &	\bf 98.50 &	\bf 73.92 &&	61.59 &	71.29 &&	25.06 &	82.50 &&	46.72 &	{\em67.01}      \\

AttNet~\cite{yu2019attributing} &67.21 &	54.76 &	65.14 &	49.35 &	50.65 &&	50.83 &	57.63 	&48.06 &&	43.83 &	47.50 &&	49.64 &	55.28 &&	50.53 &	{\em53.11}                          \\

Xception~\cite{chollet2017xception}&83.90 &	54.34 &	50.71 &	50.10 &	52.68 &&	63.55 &	50.03 &	55.73 &&	68.71 &	94.48 &&	48.21 &	58.33 &&	50.08  & {\em60.24}
 \\

DenseNet~\cite{huang2017densely}  &93.30  &	73.21  &	63.78  &	\bf 80.70  &	57.55  & &	73.66  &	86.47  &	60.72 & &97.66  &	94.45  & &	57.28  &	77.22  & &	59.94 & {\em75.07} \\

PatchForensics~\cite{chai2020makes}   &85.31  &		72.27  &\bf 71.25 &	70.35 &	\bf 69.54 &&	69.22 &	68.71 &	67.43 &&	63.47 &	61.18 &&	\bf{61.12} &	61.14 &&	60.50 &	{\em67.81}    \\

CNNDetect~\cite{wang2020cnn} & \underline{94.04}  &67.58  & 57.97 & 63.15 &	55.68 &&63.97  &73.29 &55.71  && \bf 98.32  &\underline{96.26} && 54.79  & \bf 86.67 && \bf 90.82  &{\em73.71}  \\

\midrule
\bf GFD-Net (DenseNet)  & 93.85 &	\underline{76.73} &	\underline{69.47} &	\underline{79.45} &	62.78 & &	\underline{75.85} &	\underline{96.90} &	67.73 &&	\underline{98.13} &	\bf 97.16 &&	 \underline{61.10} &	83.61 &&	87.75 &	\bf{\em 80.81} \\

\bf GFD-Net (ResNet50) & \bf 95.53 &	\bf 80.20 &	66.26 &	74.05 &	\underline{64.65} &&	\bf 78.73 &	93.40 &	\underline{67.94} &&	92.39 &	94.31 && 56.09 &	\underline{85.00} &&	\underline{90.47}& \underline{\em 79.92}  \\ 

\bottomrule
\setlength{\belowcaptionskip}{8pt}%
\end{tabular}}
\caption{Accuracy(\%) on cross-generator fake image detection. The best and second-best results in each column are {\bf boldfaced} and \underline  {underlined} respectively. In the last column, we show the averaged accuracy over all test sets.}
\label{tab:realfake} 
\end{table*}

The comparison results are listed in Table \ref{tab:att128}. In closed world testing, all methods achieve good performance ($\textgreater$ 90\% accuracy) and our method performs perfectly ($\sim$ 100\% accuracy). However, in open world testing, the performance degrades cross all methods, which demonstrates the training data of generation models largely influences the accuracy of fake image attribution. Our method achieves state-of-the-art performance in open-world testing, showing our method captures more content invariant features relating to the architecture of GAN networks.

\subsubsection {Evaluation on 1024px GANs}
\label{sec:1024}

The experiment is conducted on 4 classes: real, StyleGAN, StyleGAN2 and ProGAN. We sample 10k real images equally from FFHQ~\cite{brock2018large} and CelebAHQ, and 20k generated images equally from the public available StyleGAN-FFHQ, StyleGAN2-FFHQ and ProGAN-CelebHQ, resulting in a dataset with 20k images for each class. Then we split each set into 15k training, 1k validation, and 4k for closed world testing. To evaluate the generalization ability, we conduct open world testing on several StyleGAN and StyleGAN2 models trained on diverse datasets collected from  \footnote{\url{http://www.seeprettyface.com/mydataset.html}}.
  The open-world StyleGAN models include 10 models trained respectively on Yellow, Model, Asian stars, kids, elders, adults, people wearing glasses, male, female, and people with a smile. The open-world StyleGAN2 models include 4 models trained on Yellow, Model, Asian stars and kids. We generate 4k images from each model for testing. The generated samples are shown in the supplementary material. We apply random crop in training and center crop in testing on the images to 224 pixels. 

The results in Table \ref{tab:att1024} show that GFD-Net consistently outperforms baseline methods when tested on open-world StyleGAN and StyleGAN2 models. Specifically, the performance of PRNU, DCT and AttNet degrades largely on StyleGAN2-model and StyleGAN2-Asian-star, which may be because the datasets used to train the two models are very different from FFHQ dataset used to train the closed-world  model. However, our model can still maintain certain accuracy on these two models, showing the generalization ability of our method in open-world fake image attribution.

\subsection{Cross-Generator Fake Image Detection} 
\label{sec:general}

Direct binary fake image detection would probably fit to some explicit artifacts and thus hinder generalization. Our method can extract content irrelevant traces from fake images, which is also helpful for improving transfer performance on real/fake classification. In this section, we evaluate our method on cross-generator fake image detection as illustrated in Figure \ref{fig:exp}(b). We use {\em ForenSynths} dataset  for experiment. We train solely on ProGAN-airplane vs. LSUN-airplane and test on 13 test sets from 13 synthesis algorithms. We apply random crop in training and center crop in testing on the images to 224 pixels. No data augmentation is included for fair comparison. Two our models are trained with DenseNet and ResNet as backbone, denoted as GFD-Net(ResNet50) and GFD-Net(DenseNet)  respectively.


Table \ref{tab:realfake} summarizes the results. Comparing the results, we have following findings: 1) GFD-Net shows better transferability than baselines on the average. Although GFD-Net does not always outperforms all baselines, it usually gets a second highest accuracy which is very close to the highest, indicating that 
our method captures the common-shared fake traces on the test sets. 
2) Comparing GFD-Net(ResNet50) and CNNDetect which both use ResNet50 as the backbone, large improvements can be observed on most test sets especially on StyleGAN, CycleGAN, StarGAN, and GauGAN. Similar large improvements can also be found when comparing GFD-Net(DenseNet) and DenseNet, the accuracy on StarGAN, GauGAN and FF++ improves at least 10\%. Note that GFD-Net(ResNet50) and GFD-Net(DenseNet) only use the encoder with a classification head at inference time, which have the same architectures with ResNet50 and DenseNet. This result demonstrates that the fingerprint learning process in our model helps the generator to capture more generalized fake traces from input images and thus improves the generalization ability.

\begin{table}[t]
\begin{center}
\scalebox{0.8}{
\begin{tabular}{lcccc}
\toprule
\multirow{1}{*}{Method} & \multicolumn{1}{c}{Closed-world} & &  \multicolumn{1}{c}{Open-world} \\
\midrule
$G$ &      \bf99.99                 &  &  74.77                 &              \\
$G$+$D$ &     99.97           &   &   72.86                &              \\
$G$+$C$  &     \bf99.99             &   &  74.70                 &              \\
$G$+$D$+$C$  &     \bf 99.99             &   &  76.40                 &              \\
$G$+$D$+$C$+$L_{percept}$  &     \bf 99.99             &   &  \bf 78.72                 &              \\ \bottomrule
\end{tabular}}
\end{center}
\caption{Quantitative analysis on 128px cross-dataset fake image attribution. The best results in each column are \bf{boldfaced}. } 
\label{tab:ablation}
\end{table}

\subsection{Ablation Study}
\label{sec:ablation}

\begin{figure*}
\begin{center} 
\includegraphics[width=0.9\textwidth]{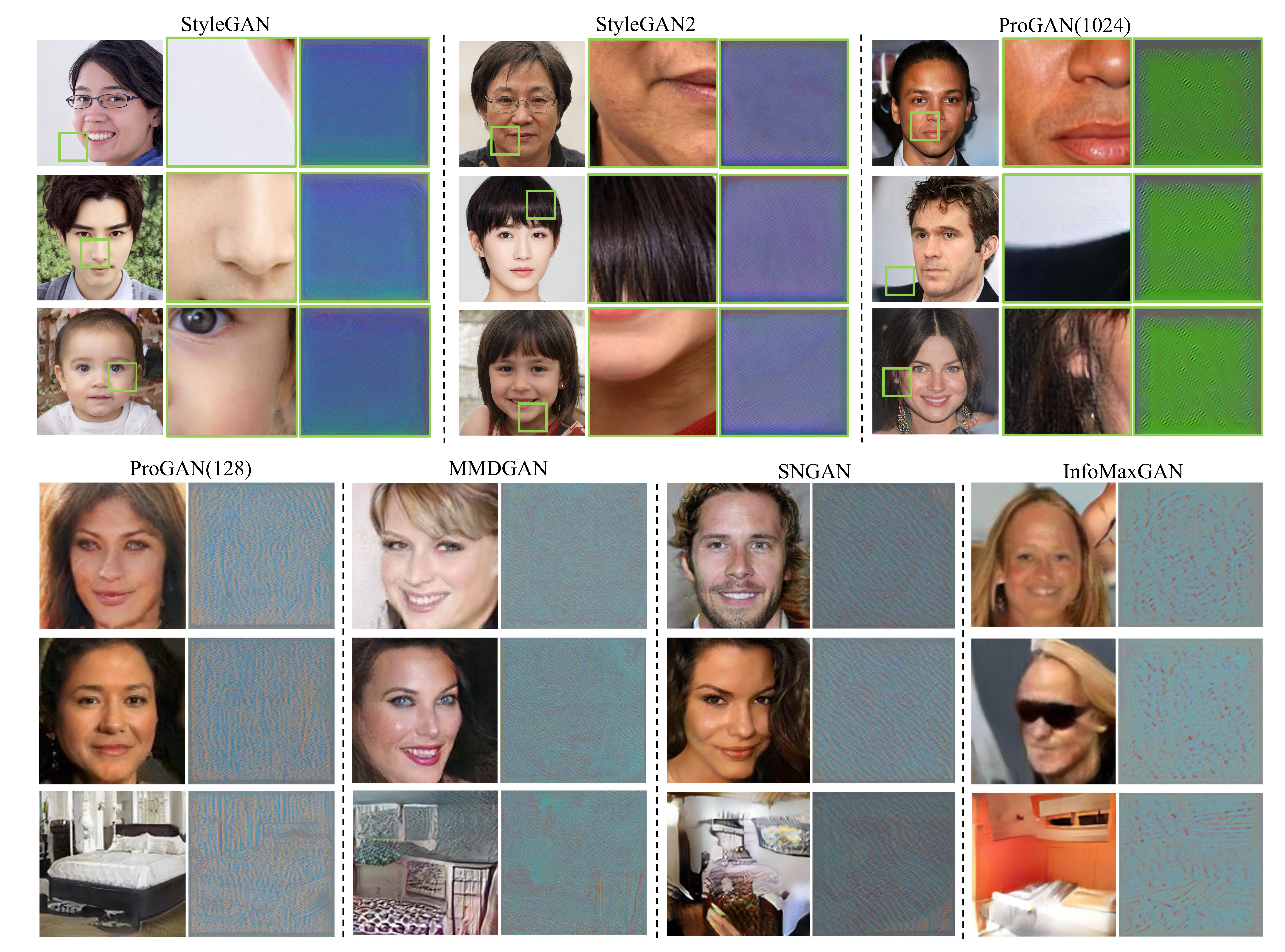}     
\end{center}  
   \caption{Generated fingerprints by our method on images from StyleGAN, StyleGAN2, ProGAN(1024), ProGAN(128), MMDGAN, SNGAN and InfoMaxGAN.}
\label{fig:finger1024}
\end{figure*}

\subsubsection{Quantitative Analysis}

To confirm the effectiveness of each component, we evaluate how generalization capacity is improved on 128px fake image attribution. Our baseline is the generator $G$ with a classification head. We compare $G$, $G$+$D$, $G$+$C$, $G$+$D$+$C$ and $G$+$D$+$C$+$L_{percept}$. The results are reported in Table \ref{tab:ablation}. In closed-world testing, all methods has good performance, the difference mainly exists in open-world testing. In open-world testing, the baseline model can achieve an accuracy of 74.77\%. Adding $C$ or $D$ alone to the baseline model doesn't improve the performance. However, with $C$ and $D$ both added, the accuracy increases from 74.77\% to 76.40\%, which demonstrates that $D$ and $C$ functions collaboratively to help improve the generalization ability. With $L_{percept}$ further added, the accuracy further increases from 76.40\% to 78.72\%, showing $L_{percept}$ also helps the generator to capture generalized content-irrelevant features.

\subsubsection{Qualitative Analysis}
\begin{figure}[t]
\begin{center}
\includegraphics[width=0.9\linewidth]{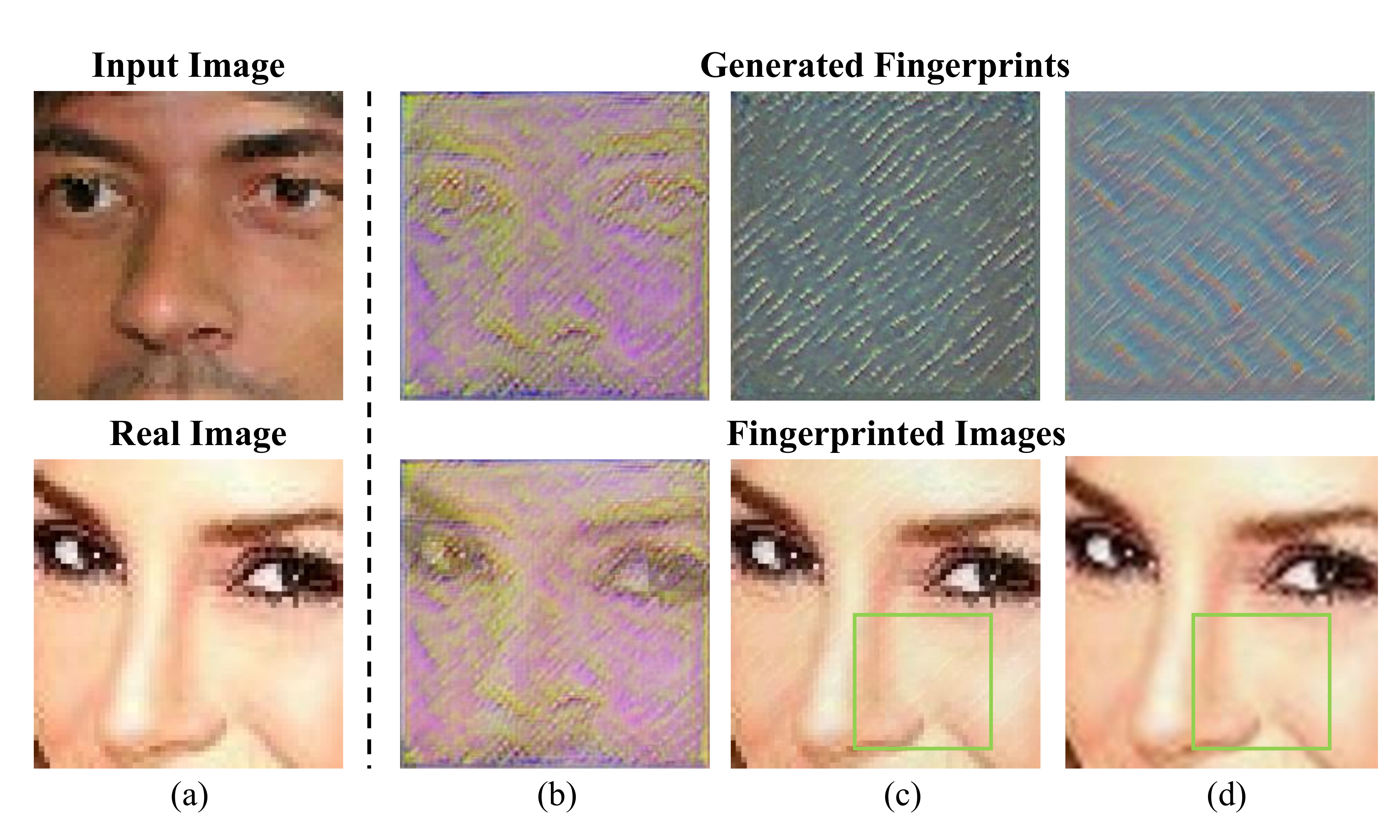} 
\end{center}
   \caption{Qualitative analysis on fingerprint generation. (a) Input image from SNGAN (top) and real image (bottom). (b)(c)(d) Generated fingerprint (top) and fingerprinted image (bottom) by $G$+$C$, $G$+$C$+$D$ and $G$+$C$+$D$+$L_{percept}$, respectively.}
\label{fig:ablation}
\end{figure}

We do qualitative analyses to demonstrate the effect of each component on fingerprint generation. In Figure \ref{fig:ablation},  we visualize the fingerprints generated by $G$+$C$, $G$+$D$+$C$ and $G$+$D$+$C$+$L_{percept}$. The first row contains an input image generated by SNGAN and extracted fingerprints from the input image by different networks. The second row contains a real image and fingerprinted images (add extracted fingerprints in the first row on the real image).  

Comparing the generation results, with auxiliary classifier $C$ only (Figure \ref{fig:ablation}b), the generator can extract a fingerprint with periodic patterns. However, the fingerprint contains much semantic information such as eyes, nose, and eyebrow, which are evidently exhibited on the fingerprinted image.
With $D$ added (Figure \ref{fig:ablation}c), the semantic content is largely suppressed and clear texture is shown on the extracted fingerprint, which demonstrates that adversarial learning helps the generator extract content-irrelevant patterns from the input image. Though the fingerprint leaves little traces on the fingerprinted image, some subtle traces can still be perceived (in the green box). With $L_{percept}$ added (Figure \ref{fig:ablation}d), the traces in the green box are eliminated, which indicates that the perceptual loss further inhibits the generator from learning semantic clues from the input image. 

\subsection{Fingerprint Analysis}
 \label{sec:finger}

{\bf \em What do GAN fingerprints look like?} Figure \ref{fig:finger1024} visualizes the fingerprints extracted from StyleGAN, StyleGAN2, ProGAN(1024), ProGAN(128), MMDGAN, SNGAN and InfoMaxGAN. Fingerprints of 1024px GAN are extracted from cropped patches. Our model successfully disentangles fingerprints from GAN images, which are common and stable among all generated by the same GAN and different between different GANs. We find that the fingerprints share similar periodic characteristics among all types of generalized images. The difference between different types mainly exists in the thickness, stretch direction, and bending of the fingerprint. We also find GAN fingerprints exist globally in images no matter in textured regions or smooth regions.

\begin{figure}[t]
\begin{center}
\includegraphics[width=1\linewidth]{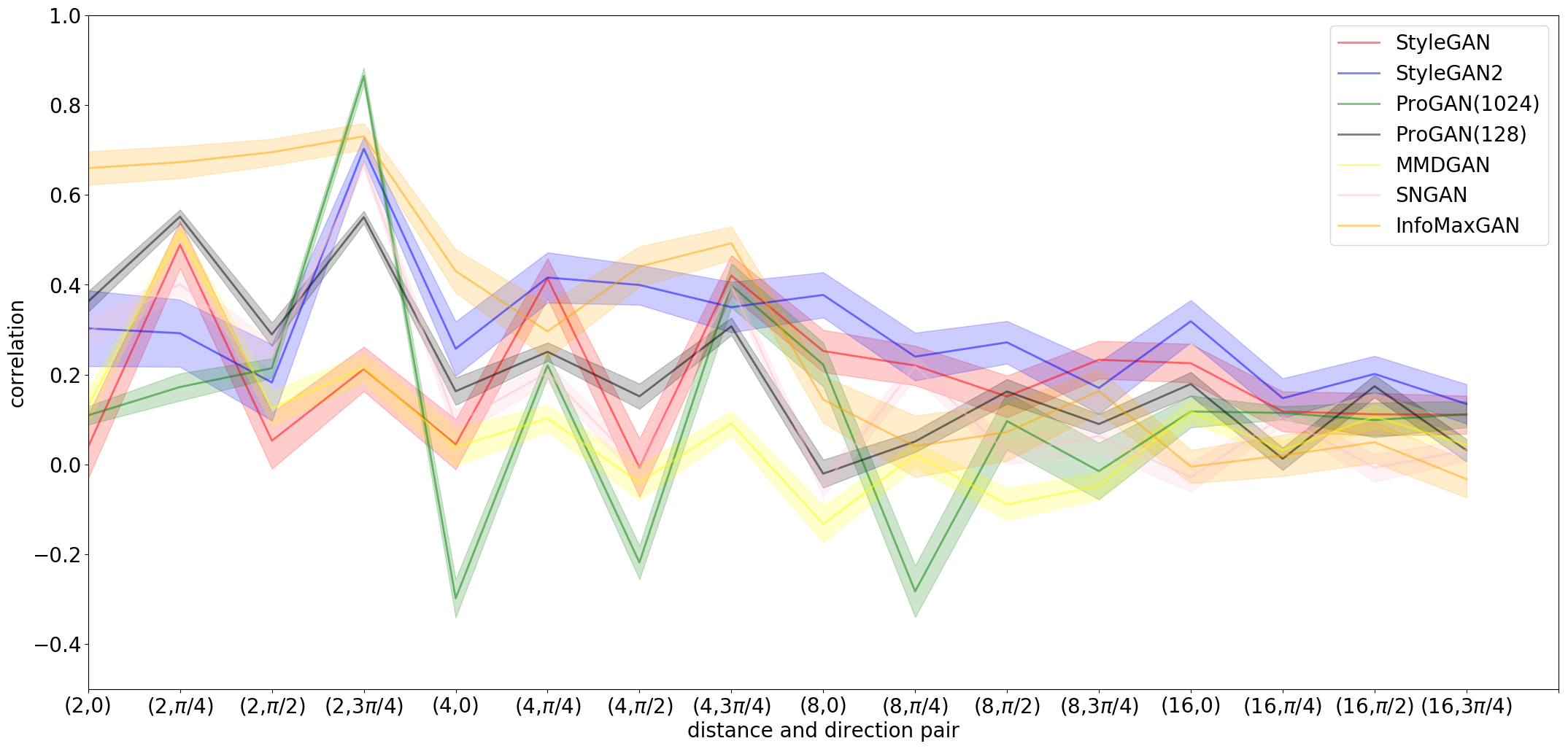} 
\end{center}
   \caption{Means and variances of GLCM correlation vectors calculated on generated images from StyleGAN, StyleGAN2, ProGAN(1024), ProGAN(128), MMDGAN, SNGAN and InfoMaxGAN.}
\label{fig:glcm}
\end{figure}

{\bf \em Qualitative analysis on GAN Fingerprints.} We calculate the gray-level co-occurrence matrix (GLCM)~\cite{haralick1973textural} from the generated fingerprints. From the GLCM, we compute texture correlation $C_{d}^{\theta}$, which measures how correlated a pixel is to its neighbor at $d$ distance offset and $\theta$ direction offset. We calculate $C_{d}^{\theta}$ on the fingerprints generated from 1000 samples for each GAN, where $d \in \{2,4,8,16\}$ and $\theta \in \{0,\pi/4,\pi/2,3\pi/4 \}$. Then for each fingerprint image, we get a $4 \times 4$ correlation matrix for every combination of $d$ and $\theta$. We reshape the matrix into a vector and calculate its mean and variance. As Figure~\ref{fig:glcm} shows, the fingerprints of different GANs have distinct correlation vectors, indicating each GAN has its specific property. The correlation is relatively larger in positions such as $(2,\pi/4)$, $(2,3\pi/4)$ and $(4,3\pi/4)$, showing a stronger correlation between adjacent pixels in $\pi/4$ and $3\pi/4$ direction.
	 
\begin{figure}[t]
\begin{center}
\includegraphics[width=0.9\linewidth]{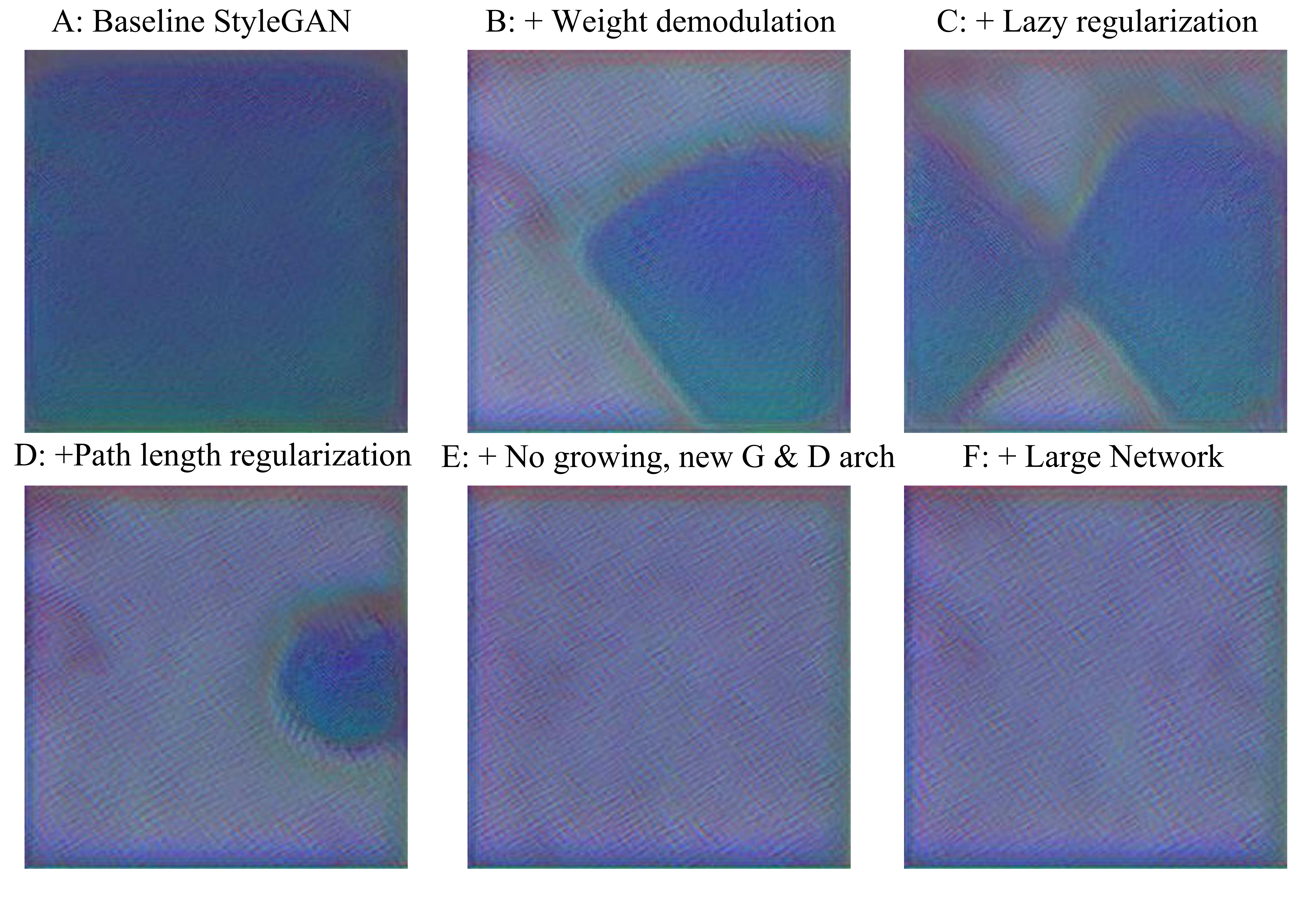}  
\end{center}
   \caption{Fingerprint extraction results of StyleGAN2 with Config A $\sim$ F.}
\label{fig:config}
\end{figure}

{\bf \em Which factors dominate GAN fingerprints?} We generate images awith the publicly available StyleGAN2 models with six configurations (Config A to F, where A and F correspond to official StyleGAN and StyleGAN2 respectively). Then we extract fingerprints from these images with our fingerprint generator. The result in Figure~\ref{fig:config} shows that: 1) The image generated by Config E model has the same fingerprint with Config F (StyleGAN2). 2) Fingerprints under Config B,C, and D appear to be a combination of the StyleGAN and StyleGAN2 fingerprint.
 Comparing these architectures, instance normalization is replaced by a demodulation operation from Config A to B, which injects StyleGAN2 fingerprint onto the image. From Config D to E,  
 the feedforward generator and discriminator are replaced by a skip generator and a residual discriminator, which results in a pure StyleGAN2 fingerprint on the image. From Config E to F, the number of feature maps is doubled, which has little influence on the fingerprint. The results demonstrate that the construction and combination of layers (replace instance normalization with demodulation operation and change feed-forward network to skip-and-residual network) have larger influence on the fingerprint, while changing feature channel number have less effect.

\section{Conclusion}
We propose GFD-Net to disentangle the fingerprint from GAN-generated images and attributing fake images to their sources simultaneously. Experiment results demonstrate the effectiveness and generalization ability of the network in not only fake image attribution but also detection. We further analyze different GAN fingerprints, showing they share similar periodic patterns and distinct in the specific textures. We also find GAN fingerprint is mostly dominated by the construction and combination of layers. We believe our work advances both fake image attribution and detection, and would bring some insights to GAN dissection.

{\small
\bibliographystyle{ieee_fullname}
\bibliography{egbib}
}

\end{document}